\documentclass[runningheads]{llncs}
\pdfoutput=1

\usepackage[utf8]{inputenc}
\usepackage{makeidx}
\usepackage{graphicx}
\usepackage{amsmath,amssymb} 
\usepackage{mathrsfs}
\usepackage{color}

\usepackage{todonotes}
\usepackage{hyperref}
\usepackage{siunitx}
\usepackage{pgfplots}
\pgfplotsset{compat=newest}
\usepgfplotslibrary{groupplots}
\usepgfplotslibrary{dateplot}

\usepackage{multirow}

\usepackage{booktabs}
\newcommand{\pcent}[1]{\SI{#1}{\percent}}
\newcommand{\etal}{\emph{et~al.}}

\begin{document}
	\pagestyle{headings}
	\mainmatter

	\title{Classification-Specific Parts for Improving Fine-Grained Visual Categorization}

	\titlerunning{Classification-Specific Parts for Improving FGVC}
	\authorrunning{D. Korsch, P. Bodesheim and J. Denzler}
	\author{
		Dimitri Korsch$^1$,
		Paul Bodesheim$^1$ and
		Joachim Denzler$^{1,2}$
	}
	\institute{
		$^1$Computer Vision Group, Friedrich-Schiller-University Jena, Germany
		\\
		$^2$Michael Stifel Center Jena, Germany
	}

	\maketitle
	\vspace{-.1cm}
\begin{abstract}
Fine-grained visual categorization is a classification task for distinguishing categories with high intra-class and small inter-class variance.
While global approaches aim at using the whole image for performing the classification, part-based solutions gather additional local information in terms of attentions or parts.
We propose a novel classification-specific part estimation that uses an initial prediction as well as back-propagation of feature importance via gradient computations in order to estimate relevant image regions.
The subsequently detected parts are then not only selected by a-posteriori classification knowledge, but also have an intrinsic spatial extent that is determined automatically.
This is in contrast to most part-based approaches and even to available ground-truth part annotations, which only provide point coordinates and no additional scale information.
We show in our experiments on various widely-used fine-grained datasets the effectiveness of the mentioned part selection method in conjunction with the extracted part features.
\end{abstract}

\section{Introduction}
\label{sec:introduction}

Fine-grained visual categorization (FGVC) is a challenging subdiscipline of computer vision and aims at distinguishing similar classes of objects that belong to a common major class like birds \cite{NABirds,CUB_200_2011}, cars \cite{StanfordCars} or flowers \cite{Flowers102}.
The latest FGVC challenges (like \cite{iNaturalist}) highlight both importance and difficulties of fine-grained categorization.
As shown by others before, a careful selection of data \cite{Cui_2018_CVPR_large} or gathering additional data from the Internet \cite{krause2016unreasonable} for the pretraining of a convolutional neural network (CNN) can yield impressive state-of-the-art results.

In general, the proposed solutions found in the literature can be divided into algorithms working with global image features~\cite{lin2015bilinear,Simon19:Implicit} and part-based or attention-based methods~\cite{Fu_2017_CVPR,Ge19Weakly,he2019and,zhang2019unsupervised,zheng2017learning}.
From the empirical results reported in these works, it is difficult to conclude which basic approach (global or part-based) works best, since both are competitive in terms of recognition performance.
Due to the fact that categories of fine-grained recognition tasks often differ only in small details, part-based features that consider local image regions seem to be promising because they are able to explicitly focus on such distinct patterns.
For example, if two bird species can only be distinguished by a characteristic spot on the head, a part feature representing the image region covered by the head of the bird would be beneficial for separating these two classes.
Furthermore, parts can support the classification in case of only few training samples or highly imbalanced class distributions in the training set, e.g., by applying transfer learning techniques~\cite{Goering14:NPT}.
For analyzing classification results and failure cases, the attribution of classifier decisions to features and relevant image regions is an important step and parts are helpful for detailed investigations in order to gain a better understanding of the specific recognition task and the problem domain.

\begin{figure}[t]
	\centering
	\includegraphics[width=\textwidth]{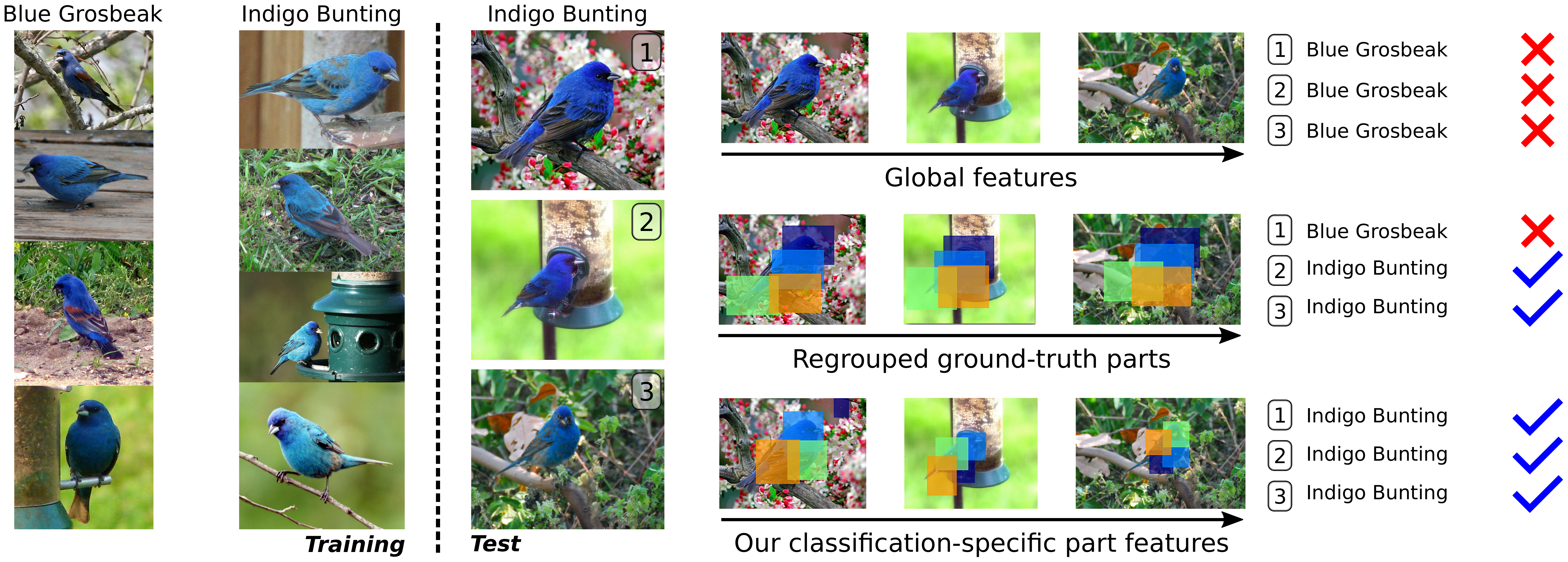}
	\caption{An example from our experiments as a motivation for our approach: two visually similar classes are confused by a baseline classifier with global features.
	A part-based classifier using ground-truth annotations for anatomical parts is able to correct some of the predictions due to additional local information.
	However, our parts are estimated using a-posteriori classification knowledge and therefore focus on distinguishing highly similar classes.
	This allows for resolving misclassifications with the additional benefits of automatically determining the spatial extent of each part and being independent from manual annotations.}
	\label{fig:teaser}
\end{figure}

If ground-truth~(GT) annotations for part locations are provided, they are usually referring to an underlying concept, e.g., the anatomical parts of a bird such as head, beak, belly, wings, and legs.
Although being plausible from a human perspective, these parts may not be the best choice for achieving the highest classification accuracy with a machine learning model.
In addition, not all annotated parts are equally relevant for every test image and it can be shown that an optimal part selection would lead to superior performance compared to state-of-the-art methods using all available parts~\cite{Korsch18:Defense}.
Especially in case of noise, few characteristic parts can be outweighed by the remaining larger set of irrelevant parts that confuse the classifier and lead to misclassifications.
In our experiments, we have observed that quality of parts is more important for an improved classification accuracy than quantity.
For example, if we regroup the provided ground-truth parts for the CUB-200-2011 birds dataset~\cite{CUB_200_2011} in more coarse but also more distinct parts, namely ``head'', ``body'', ``legs'', and ``tail'', the recognition performance can be enhanced.

However, since ground-truth annotations are not available for all applications and manual part annotations are expensive, an efficient and robust part detector is required.
Such a detector has to deal with the following two main questions.
First, what are the interesting and important locations that enhance the classification performance?
Second, given certain part location, how and to which extent should the part features be extracted?

In this work, we tackle those questions and show that our classification-specific part estimation is able to improve classification accuracies on various fine-grained datasets.
By studying failure cases of a baseline classifier that makes use of either global image features or part-based features extracted from available ground-truth annotations of the CUB-200-2011 birds dataset~\cite{CUB_200_2011}, we have observed that many class confusions occur between visually very similar classes (more details in Sect.~\ref{sec:methods} and Fig.~\ref{fig:cub200_conf_matrix}).
Hence, the idea of our approach is to identify relevant parts based on an initial classification with global features.
By considering only the most important features for this initial decision, we estimate parts that are likely to be relevant for visually similar classes as well.
With these new parts that are specifically estimated for a given test image based on additional knowledge from an initial classification, we aim at resolving misclassifications between very similar classes.
This scenario is also visualized in Fig.~\ref{fig:teaser}.

Our proposed approach consists of the following steps.
First, we perform a feature selection in order to estimate the most important features for the current classification task using a baseline classifier with global image features.
The idea is then to estimate the most important regions in the image with respect to the actual classification task by only taking the most important features for the part localization into account.
From these regions, we estimate parts as bounding boxes with automatically determining the spatial extent (scale) of each part.
This is an advantage over most part-based approaches and ground-truth annotations.
These provide only $x$ and $y$ coordinates of the part locations such that the size of each part has to be selected appropriately (and is usually fixed for all parts and all images).
Given the newly estimated parts, we extract features for these parts as a rich representation that can then be used to improve the classification.
More details are given in Sect.~\ref{sec:methods}.

As it is common practice for many computer vision applications nowadays, our classification scheme relies on features computed with a CNN (called CNN features).
Such features can easily be computed by applying either a pretrained CNN model as it is or a pretrained model that has been fine-tuned on the training set of the desired application.
In our experiments (Sect.~\ref{sec:experiments}) we show that our approach: (i)~improves the performance of the baseline methods, (ii)~is competitive with other part-based classification approaches, and (iii)~achieves state-of-the-art accuracies in some applications.

\section{Related Work}
\label{sec:related_work}

Fine-grained visual categorization is a challenging and non-trivial classification task.
Hence, there are diverse ways to tackle the problem.
On the one hand, there are approaches that only use the global information of the image.
The idea is either to use a clever way of pretraining the classifier or to use different feature pooling strategies.
On the other hand, part-based approaches are applied which differ in the various part detection and extraction techniques.

\subsection{Global Feature Representations}
\label{sub:fine_grained_visual_categorization}

First, we consider approaches using only the global information of the image.
Cui \etal~\cite{Cui_2018_CVPR_large} use a smart strategy in order to pretrain a CNN by taking large-scale datasets like ImageNet or iNaturalist into account, which offer a lot of data.
Unfortunately, the difference between these datasets and the desired fine-grained datasets is too big.
Hence, they suggest to preselect certain classes from the large-scale datasets which match best to the fine-grained training images and show that this preselection improves the performance drastically.

Krause \etal~\cite{krause2016unreasonable} enrich the training data with images from the Internet.
They use Google Image Search in order to gather additional images for every training class.
Though, the retrieved samples may not belong to the queried class, they show that even this noisy data improves the recognition performance by a large amount.
Nevertheless, they cannot ensure that the collected data does not contain some images from the validation or the test set, since all these images are also publicly available.
Hence, although an impressive effectiveness of noisy data is shown, one should look at the reported results with caution.

Other approaches focus on advanced ways of feature pooling.
Here, bilinear pooling by Lin \etal~\cite{lin2015bilinear} or more general the alpha-pooling by Simon \etal~\cite{Simon19:Implicit} are the most common techniques.
Their aim is to highlight features that may have a greater impact on the classification task.

\subsection{Part-based Recognition Approaches}
\label{sub:part_based_recognition}

The second main direction for tackling fine-grained recognition tasks consists of methods that rely on part-based representations.
A straightforward way of implementing a part-based recognition system is to employ the ground-truth part annotations if they exist (e.g., for the CUB-200-2011 birds dataset \cite{CUB_200_2011}).
Since these annotations are expensive and most fine-grained datasets do not provide them, weakly supervised part detectors are a common choice \cite{Fu_2017_CVPR,he2019and,Simon_2015_ICCV,zheng2017learning}.
The only supervision that these detectors use are class label annotations.

Fu \etal~\cite{Fu_2017_CVPR} and Zheng \etal~\cite{zheng2017learning} present similar approaches to extend CNNs with attention networks.
The first work considers adjusting the attention recurrently and extracting additional information defined by the attention on different scales.
On the other hand, the later work extracts multiple attentions in a single step.
The extraction is done by localizing interesting areas from feature maps, regrouping them, and using these grouped areas as parts.
In both cases, the whole system is trained end-to-end.

He \etal~\cite{he2019and} propose a sophisticated reinforcement learning method in order to estimate how many and which image regions are helpful to distinguish the categories.
They use multi-scale image representations in order to localize the object and then estimate discriminative part regions.

Simon \etal~\cite{Simon_2015_ICCV} identify part proposals with the aid of back-propagation.
Afterwards, these proposals are used to fit a constellation model that determines which of the proposals are more likely to identify real parts.
The part proposals with the highest match are then used to extract part features on different scales.

\section{Classification-Specific Part Estimation}
\label{sec:methods}

\begin{figure}[t]
	\centering
	\includegraphics[width=0.7\textwidth]{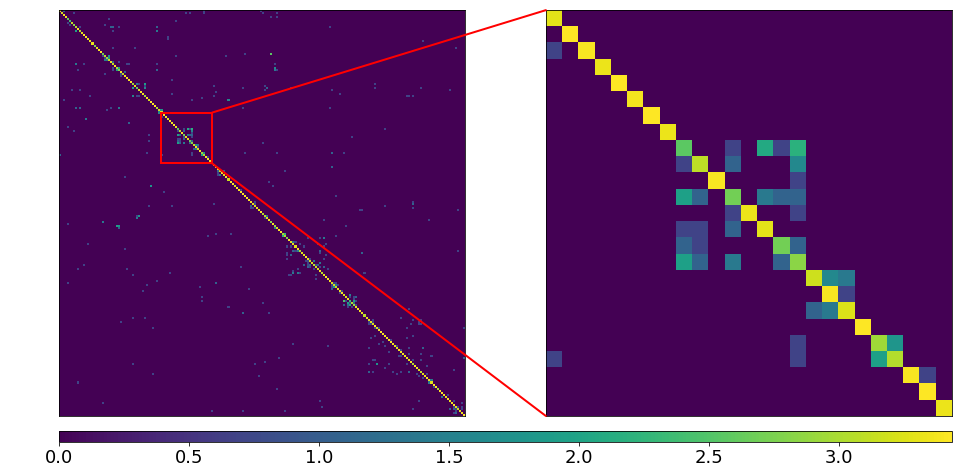}
	\caption{
	Confusion matrix created from CUB-200-2011 predictions using global features only.
	The values are absolute number of correct predictions in log-scale.
	Similar classes have consecutive class indexes and are confused more often.
	Here you can see the classes 59-66, which are different gull species, e.g. California Gull, Herring Gull, Ivory Gull or Western Gull}
	\label{fig:cub200_conf_matrix}
\end{figure}

In this section we describe our classification-specific part estimation approach that makes use of an initial classification based on global image features.
The goal is to estimate parts depending on this first (and probably wrong) decision, such that these parts can help to spot the tiny details.
These details are important for distinguishing visually similar classes in order to either confirm an initially correct classification based on the specific part features or to correct an initially wrong classification due to an enhanced representation of the important parts only.
Here, we assume that many confusions of a classifier based on the global image features occur between visually very similar classes and that in those cases, the small details that are characteristic for distinguishing them are not well represented by the global features (which in general have to work for distinguishing all classes).
Fig.~\ref{fig:cub200_conf_matrix} visualizes this confusion and confirms our assumption.
Hence, we look for the important image regions that have led to the initial classification.
Then we derive new parts from these regions under the assumption that the resulting parts are also more relevant for disentangling visually similar classes.
Our estimated parts are therefore classification-specific rather than based on human knowledge, e.g., from an anatomical point of view in case of the birds.

\begin{figure}[t]
	\centering
	\includegraphics[width=\textwidth]{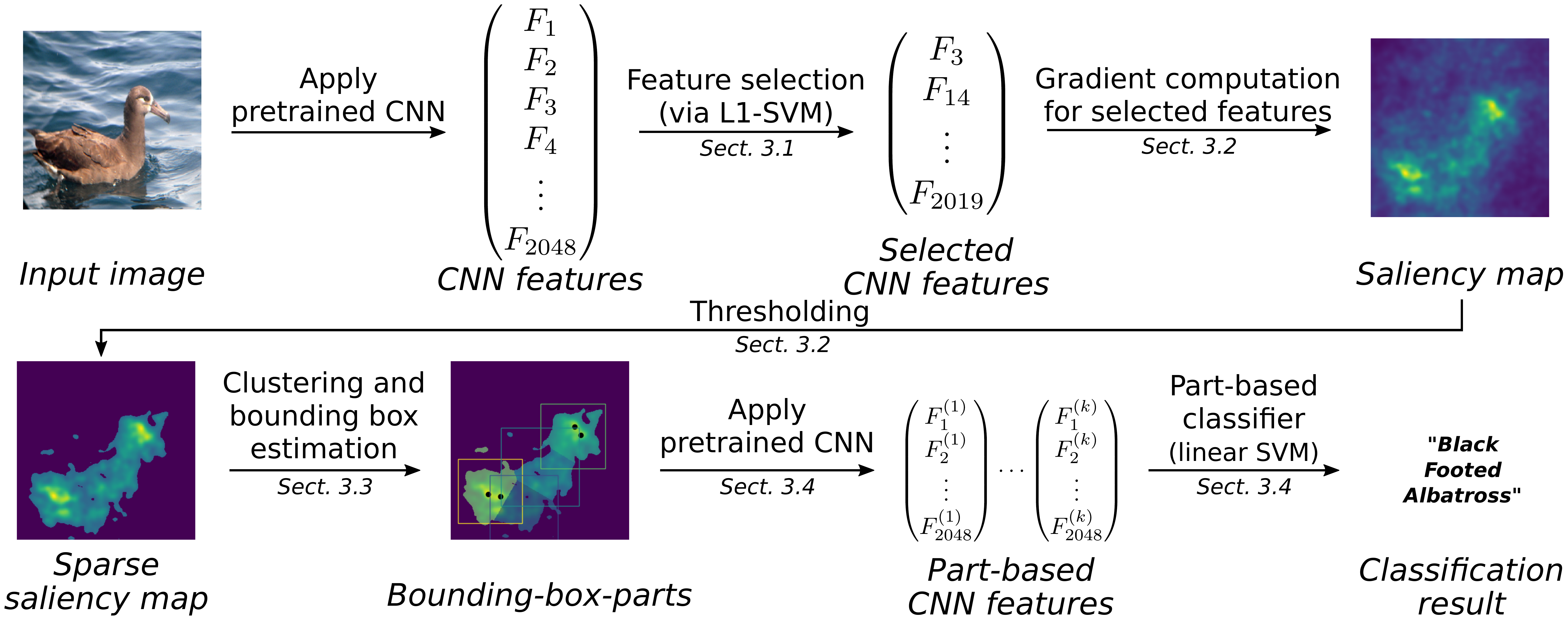}
	\caption{The pipeline for our classification-specific part estimation.}
	\label{fig:pipeline}
\end{figure}

The pipeline we propose in this paper is visualized in Fig.~\ref{fig:pipeline} and will be outlined in the following.
First, we describe the feature selection method (Sect.~\ref{sub:feature_selection}).
Next, based on these selected features, we illustrate how relevant pixels and image regions are identified as candidates for the part locations (Sect.~\ref{sub:part_candidates}).
Third, we explain our algorithm for estimating bounding-box-parts with the advantage of automatically determining the scale of each part (Sect.~\ref{sub:bb_estimation}).
Finally, an overview about the part feature extraction from the classification-specific bounding-box-parts and about the part-based classification is given in Sect.~\ref{sub:part_feature_extraction}.

\subsection{Feature Selection}
\label{sub:feature_selection}
\vspace{-.1cm}

Nowadays, a common approach in computer vision tasks is to use a pretrained neural network like a CNN, fine-tune its parameters on a dataset for the desired application, and extract features by concatenating the outputs of its penultimate layer in order to obtain a high-level descriptions of the image content.
Our approach relies on those CNN features, which typically results in high-dimensional feature vectors ($D=2048$ in our case).
In case of a fine-grained recognition task, the recognition system often has to focus on some specific information within the features in order to spot tiny details that distinguish two similar classes.

Therefore, we first perform a feature selection in order to estimate the most important features for the current classification task.
This is done by utilizing a sparsity-inducing classifier equipped with L1-regularization, which could be either a corresponding classification layer in a CNN that allows for end-to-end learning or an L1-regularized linear SVM classifier for the CNN features.
Optimization with L1-regularization forces the classifier decisions to be performed on only a small subset of the CNN features.
In our experiments, we tried both and found empirically that an SVM performs better in terms of recognition accuracy while still being fast during learning due to efficient SVM solvers in standard libraries like liblinear \cite{fan2008liblinear}.

In the end, our feature selection is classification-based and determines relevant features for the underlying task by optimizing feature weights during learning of SVM classifiers.
Since we consider multi-class recognition scenarios, we train a separate classifier for each class using the ``one-vs-rest'' strategy.
As the result, we obtain a subset of relevant features for each class that best distinguishes this class from all the other classes by only selecting features with nonzero weights.

\subsection{Identifying Relevant Pixels and Image Regions}
\label{sub:part_candidates}

The main idea within our part detection approach is to estimate the most important regions in the image with respect to the actual classification task.
Hence, the part localization should only take those important features into account, which have been computed with the classification-based feature selection from the previous section.
To this end, we use gradient maps \cite{simonyan2013deep} to identify the most relevant pixels in the image (indicated by large gradients), which have the largest influence on the feature extraction from the CNN model.
By restricting the gradient map computations to only the previously selected subset of features, regions with large gradients are more adjusted to the classification task compared to propagating back the gradients of all features to the input image.
Since our feature selection is based on a multi-class one-vs-rest SVM classifier, only selected features of the class assigned by this classifier are used for the gradient map computations.
Thus, we incorporate knowledge of a baseline classifier with global features in our part detection algorithm.

The gradient maps are treated as saliency maps in our approach in order to guide the part detection and they depend on the used CNN features.
In case of many currently used CNN architectures (Inception \cite{Szegedy_2016_CVPR}, ResNet \cite{he2016resnet}, etc.), features are computed by averaging the values within each of the $D$ output channel of the last convolutional layer.
Typically, these output channels are called feature maps and the aforementioned average pooling results in a single number for each feature map.
Given $D$ feature maps $\vec{F}^{(1)}(\vec{I}),~\ldots,~\vec{F}^{(D)}(\vec{I})$ of size $s \times u$ for an image $\vec{I}$, this pooling step for computing the elements $f^{(d)}(\vec{I})$ of the $D$-dimensional feature vector $\vec{f}(\vec{I})$ can be written as follows:
\vspace{-.2cm}
\begin{equation}
f^{(d)}(\vec{I}) = \dfrac{1}{s \cdot u}\sum_{j=1}^s \sum_{j'=1}^u F_{j,j'}^{(d)}(\vec{I})\quad\forall\ d\in\{1,\dots,D\} \; .
\end{equation}
Consequently, each value in a feature vector corresponds to a single feature map and since the feature selection method mentioned in Sect.~\ref{sub:feature_selection} is applied to the feature vectors, it can also be viewed as applying the feature selection to the feature maps, i.e., the output channels of the last convolutional layer.

Like Simonyan \etal~\cite{simonyan2013deep} and Simon \etal~\cite{Simon_2015_ICCV}, we use back-propagation through the CNN to identify the regions of interest for each selected feature map.
Based on the feature map subset $\mathfrak{D} \subset \{1, \dots, D\}$ chosen by the feature selection from Sect.~\ref{sub:feature_selection}, we compute a saliency map $\vec{M}(\vec{I})$ for an image $\vec{I}$ as follows:

\vspace{-.2cm}
\begin{equation}
	\label{eq:saliency_sum}
	M_{x,y}(\vec{I})
	  = \dfrac{1}{|\mathfrak{D}|} \sum_{d \in \mathfrak{D}} \left| \dfrac{\partial}{\partial I_{x,y}} f^{(d)}(\vec{I}) \right|
	  = \dfrac{1}{|\mathfrak{D}|} \sum_{d \in \mathfrak{D}} \left| \dfrac{\partial}{\partial I_{x,y}} \dfrac{1}{s \cdot u}\sum_{j=1}^s \sum_{j'=1}^u F_{j,j'}^{(d)}(\vec{I}) \right| \;  .
\end{equation}

After estimating the saliency maps, we normalize the resulting values to the range $[0 \dots 1]$ and determine a threshold to discard pixels and regions of low saliency at an early stage.
We use the mean saliency value as a threshold.
We have also tested Otsu's thresholding method~\cite{otsu1979threshold} and it achieved similar performance.
The resulting sparse saliency map, which now contains only pixels with large saliency values, is used in the next step for estimating location and spatial extent of parts.

\subsection{Estimating Bounding-Box-Parts}
\label{sub:bb_estimation}

Given an image and a sparse saliency map that discards pixels with low saliency, a set of $k$ peaks $P = \{p_1, \dots, p_k\}$ with largest saliency can be computed using non-maximum suppression.
Each peak serves as the initialization for a new part location.
We then determine a region of high saliency around each peak, which directly defines the spatial extent of the estimated part.
Like Zhang \etal~\cite{zhang2019unsupervised}, we achieve this by \mbox{$k$-means} clustering of pixel coordinates $(x, y)$ and the saliencies $M_{x,y}$ (Eq.~\ref{eq:saliency_sum}).
Additionally, we also consider the RGB values at the corresponding positions in the input image.
The clusters are initialized with the previously determined peaks $p_1, \dots, p_k$.

This has the effect that the number of selected peaks determines the number of clusters and hence the number of parts to detect.
Second, since the peaks are sorted by their saliency values, the most important part is identified by the first cluster.
Afterwards, it is easy to translate the clusters into bounding boxes for the parts.
For each cluster we estimate the upper left and lower right corners in order to maximize the recall of the cluster pixels surrounded by the corners.
The motivation behind the recall maximization is to get bounding boxes that contain as few false negatives as possible.

The resulting bounding boxes serve as parts for the following part-based classification with the advantage that we automatically determine the spatial extent (scale) of the parts by inferring the size of the bounding boxes based on the clustering and the regression.
In contrast to this, most approaches estimate only $x$ and $y$ coordinates of the part locations such that the size of each part has to be selected appropriately.
The same holds for the ground-truth annotations of many fine-grained datasets \cite{NABirds,CUB_200_2011}.
In most cases, the size for all parts of an image is fixed, which is obviously not very suitable since parts often have different extents in the image, e.g., consider bird parts that correspond to an eye and a wing.

With our part estimation strategy, we are able to automatically determine different sizes for different parts depending on the content of the image.
Hence, we want to emphasize again that we treat parts as bounding boxes with estimated position \emph{and} estimated spatial extent in our framework rather than only considering point locations with fixed extent.

\subsection{Part Feature Extraction and Part-based Classification}
\label{sub:part_feature_extraction}

After we have estimated the bounding boxes around the maximum peaks of the sparse saliency map for each image, we extract CNN features from these bounding boxes.
This is achieved by treating each bounding box as a single image that is then processed by a pretrained CNN to extract meaningful features from the penultimate layer.
Note that this could even be the same CNN that was initially used to extract global image features for the part localization and we use the same CNN architecture for both steps.
The resulting part features and the global features are then concatenated prior to applying a linear SVM classifier.
This classifier has been trained using part features of the training images that have been computed with our part estimation approach described before.

To summarize, our part descriptors are classification-specific in the sense that we estimate location and spatial extent of parts via bounding boxes based on an initial classifier decision with its involved feature selection, i.e., our estimated parts focus on the important aspects that are relevant for the classification.

\section{Experiments}
\label{sec:experiments}

\subsection{Datasets and Implementation}
\label{sub:datasets}
\subsubsection{Datasets}
All of the experiments are performed on widely used fine-grained datasets.
These datasets belong to a single common domain (birds, cars, flowers, etc.).
Though, some of these datasets provide additional part or bounding box annotations besides the class annotations, we use only the class labels in our experiments.
A short description of these datasets can be found in the following.

\emph{CUB-200-2011} \cite{CUB_200_2011} consists of \num{5994} training and \num{5794} test images from \num{200} different bird species.
Besides the class labels, this dataset provides bounding box and part annotations.

\emph{NA-Birds} \cite{NABirds} is similar to the CUB-200-2011 dataset.
It provides besides class annotations also ground-truth part annotations.
This dataset is more challenging, since it has \num{555} classes spread over \num{23929} training and \num{24633} test images.
Although there are more training samples, the training set is not as balanced as the training set of the CUB-200-2011 dataset.
Additionally, this dataset provides a hierarchy information about the classes.

\emph{Stanford-Cars} \cite{StanfordCars} contains \num{8144} training and \num{8041} test images for \num{196} car models.
This dataset provides only bounding box annotations.

\emph{Flowers-102} \cite{Flowers102} has \num{102} different flower species spread over \num{2040} training and \num{6149} test images. Class labels are the only provided annotations.

\subsubsection{Implementation}
\label{ssub:implementation}
As backbone for our method, we use the ResNet-50~\cite{he2016resnet} CNN architecture for Stanford-Cars and Inception-V3 CNN architecture~\cite{Szegedy_2016_CVPR} for the other datasets.
For different datasets we use CNN weights proposed by Cui \etal~\cite{Cui_2018_CVPR_large}.
These weights are pretrained on either the ImageNet or the iNaturalist 2017 dataset.
To allow for fair comparisons with the recognition performances mentioned in~\cite{Cui_2018_CVPR_large}, we use ImageNet weights for Stanford-Cars and iNaturalist weights for all other datasets.
This separation makes sense, since iNaturalist consists of living things only, which matches the datasets of flowers and birds best.
On the other hand, ImageNet classes are more variable and contain also objects and vehicles, which is more suitable for a dataset of car images.
For every fine-grained dataset, we fine-tune a CNN on the corresponding training set, perform the feature selection, part localization and part extraction.
Finally, the extracted part features and the global feature are concatenated and a linear SVM classifier is trained.
In order to match the number of regrouped ground-truth parts for the CUB-200-2011 dataset mentioned in the introduction, we use $k=4$ in the part localization step, which results in four parts.

\subsection{Results}
\label{sub:results}

\subsubsection{Feature Selection Evaluation}
\label{ssub:feature_selection_eval}
First, we show that the usage of a classification-specific feature selection improves the quality of the extracted parts.
For this experiment, we first compute the gradients from the entire feature vector with respect to the input image.
Based on this gradient, we detect parts and extract features as mentioned before in Sections~\ref{sub:part_candidates},~\ref{sub:bb_estimation}, and~\ref{sub:part_feature_extraction}.
The results obtained with these features can then be compared to the results of our approach.
Although the features derived from gradients of the entire feature vector improve the recognition performance compared to the linear classification baseline, using feature selection as presented in Sect.~\ref{sub:feature_selection} yields a larger improvement, as shown in \autoref{tab:feature_selection_eval}.
We observe that the feature selection is an important ingredient in our approach.
The additional information, in form of the part features determined from the gradients of the entire feature vector, improves the classification performance.
Nevertheless, the benefits of the feature selection indicate that this additional information should be picked with care.


\begin{table}[t]
\caption{Comparison of our part extraction algorithm with and without our proposed feature selection method (\textbf{bold}~=~best per dataset).}
\begin{center}
\begin{tabular}{l@{\hspace{.2cm}}l@{\hspace{.2cm}}c@{\hspace{.2cm}}c@{\hspace{.2cm}}c@{\hspace{.2cm}}c}
\toprule
	 & & \scriptsize CUB-200-2011 & \scriptsize NA-Birds & \scriptsize Flowers-102 & \scriptsize Stanford-Cars \\
\midrule
	\multicolumn{2}{l}{Global features (baseline)}
		& 88.5 & 87.5 & \textbf{97.8} & 91.5 \\[.1cm]
	\multicolumn{2}{l}{Our parts}
		&  &  &  &  \\
	& no feature selection
		& 89.1 & 88.4 & 97.0 & 92.2 \\
	& with feature selection (Sect.~\ref{sub:feature_selection})
		& \textbf{89.5} & \textbf{88.5} & 96.9 & \textbf{92.5} \\
\bottomrule
\end{tabular}
\label{tab:feature_selection_eval}
\end{center}
\vspace{-0.6cm}
\end{table}

The saliency maps that determine the parts are computed by the sum of the gradients of every single CNN feature with respect to the input image (Eq. \ref{eq:saliency_sum}).
Hence, the feature selection reduces the summation to selected gradients only.
This means that in the experiment with feature selection, we use less information but this information is more precise which results in better classification performance.
These findings hold for all presented datasets except for the flowers dataset.
In case of flowers we see that the baseline linear classifier performs best.
One possible explanation is overfitting to the training data.
Compared to other datasets, there are on average only \num{20} training samples per class.
The other datasets contain an average of \num{30} to \num{40} samples per class.

\begin{figure}[t]
	\centering

\begin{tikzpicture}

\definecolor{color0}{rgb}{1,0.498039215686275,0.0549019607843137}

\begin{axis}[
width=.9\textwidth,
height=4cm,
tick align=outside,
tick pos=left,
x grid style={white!69.01960784313725!black},
xmin=0.5, xmax=4.5,
xtick style={color=black},
xtick={1,2},
xticklabels={CUB-200-2011,NA-Birds},
y grid style={white!69.01960784313725!black},
ylabel={Percentage},
ymin=0, ymax=2.69775390625,
ytick style={color=black},
ytick={0,0.5,1,1.5,2,2.5,3},
yticklabels={0.00\%,0.50\%,1.00\%,1.50\%,2.00\%,2.50\%,3.00\%}
]
\addplot [black]
table {%
0.775 0.634765625
1.225 0.634765625
1.225 0.748725850201283
1.225 0.78125
1.225 0.813774149798717
1.225 0.927734375
0.775 0.927734375
0.775 0.813774149798717
0.775 0.78125
0.775 0.748725850201283
0.775 0.634765625
};
\addplot [black]
table {%
1 0.634765625
1 0.390625
};
\addplot [black]
table {%
1 0.927734375
1 1.318359375
};
\addplot [black]
table {%
0.8875 0.390625
1.1125 0.390625
};
\addplot [black]
table {%
0.8875 1.318359375
1.1125 1.318359375
};
\addplot [black]
table {%
1.775 1.318359375
2.225 1.318359375
2.225 1.52670553351162
2.225 1.5625
2.225 1.59829446648838
2.225 1.85546875
1.775 1.85546875
1.775 1.59829446648838
1.775 1.5625
1.775 1.52670553351162
1.775 1.318359375
};
\addplot [black]
table {%
2 1.318359375
2 0.5859375
};
\addplot [black]
table {%
2 1.85546875
2 2.587890625
};
\addplot [black]
table {%
1.8875 0.5859375
2.1125 0.5859375
};
\addplot [black]
table {%
1.8875 2.587890625
2.1125 2.587890625
};
\addplot [black]
table {%
2.775 1.07421875
3.225 1.07421875
3.225 1.18439329521588
3.225 1.2451171875
3.225 1.30584107978412
3.225 1.46484375
2.775 1.46484375
2.775 1.30584107978412
2.775 1.2451171875
2.775 1.18439329521588
2.775 1.07421875
};
\addplot [black]
table {%
3 1.07421875
3 0.732421875
};
\addplot [black]
table {%
3 1.46484375
3 2.001953125
};
\addplot [black]
table {%
2.8875 0.732421875
3.1125 0.732421875
};
\addplot [black]
table {%
2.8875 2.001953125
3.1125 2.001953125
};
\addplot [black]
table {%
3.775 0.68359375
4.225 0.68359375
4.225 0.753871372767857
4.225 0.78125
4.225 0.808628627232143
4.225 0.927734375
3.775 0.927734375
3.775 0.808628627232143
3.775 0.78125
3.775 0.753871372767857
3.775 0.68359375
};
\addplot [black]
table {%
4 0.68359375
4 0.439453125
};
\addplot [black]
table {%
4 0.927734375
4 1.26953125
};
\addplot [black]
table {%
3.8875 0.439453125
4.1125 0.439453125
};
\addplot [black]
table {%
3.8875 1.26953125
4.1125 1.26953125
};
\addplot [color0]
table {%
0.775 0.78125
1.225 0.78125
};
\addplot [color0]
table {%
1.775 1.5625
2.225 1.5625
};
\addplot [color0]
table {%
2.775 1.2451171875
3.225 1.2451171875
};
\addplot [color0]
table {%
3.775 0.78125
4.225 0.78125
};
\end{axis}

\begin{axis}[
width=.9\textwidth,
height=4cm,
axis y line=right,
tick align=outside,
x grid style={white!69.01960784313725!black},
xlabel={},
xmin=0.5, xmax=4.5,
xtick pos=left,
xtick style={color=black},
xtick={3,4},
xticklabels={Flowers-102,Stanford-Cars},
y grid style={white!69.01960784313725!black},
ylabel={Absolute values},
ymin=0, ymax=1,
ytick pos=right,
ytick style={color=black},
ytick={0,0.2,0.4,0.6,0.8,1},
yticklabels={0,10,20,30,40,50}
]
\end{axis}

\end{tikzpicture}
	\caption{Since we perform multi-class classification with the ``one-vs-rest'' strategy, we obtain for each class a vector of sparse weights for the linear SVM due to the L1 regularization. The distribution of the number of nonzero weights over the different classes (different ``one-vs-rest'' models) is shown for each dataset that is used in our experiments. Note that both relative and absolute quantities are shown (for \num{2048} features in total), which correspond to the number of selected features.}
	\label{fig:sparsity}
\end{figure}
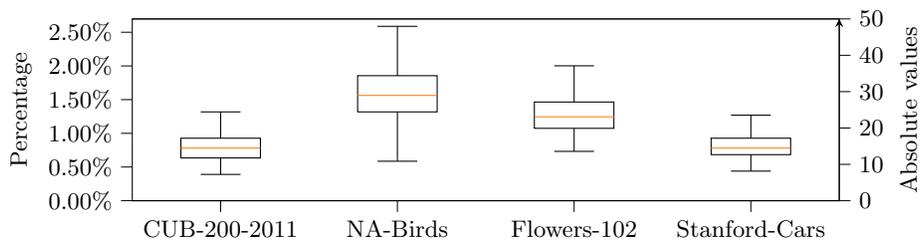

Furthermore, as Fig.~\ref{fig:sparsity} shows, the number of selected features by a L1-regularized linear classifier is beneath \pcent{3} for used datasets.
As a consequence, the number of aggregated gradients (Eq. \ref{eq:saliency_sum}) is also beneath \pcent{3}.
This fact and the results from \autoref{tab:feature_selection_eval} confirm the assumption, that quality of selected information is more important than the quantity.

\vspace{-.5\baselineskip}
\subsubsection{Part Feature Evaluation}


\begin{table}[t]
\caption{Evaluation of the extracted parts on the CUB-200-2011 dataset. Note that we have used the same CNN to extract features from the different part locations: ground-truth (GT), regrouped GT, NAC parts \cite{Simon_2015_ICCV}, and our classification-specific parts (\textbf{bold}~=~best, \textit{italic}~=~best without GT annotations).}
\begin{center}
\begin{tabular}{l@{\hspace{1cm}}c@{\hspace{.5cm}}c@{\hspace{.5cm}}c@{\hspace{.5cm}}c@{\hspace{.5cm}}c}
\toprule
    & global & parts & parts + & $\#$ of \\
	& features& only & global & parts \\ 
	\midrule
	Global features (baseline)
		& 88.5 & - & - & - \\[.2cm]
	GT parts
		& - & 87.9 & 89.8 & 15  \\
	Regrouped GT parts
		& - & 86.9 & \textbf{90.2} & 4 \\[.2cm]
	NAC part locations of \cite{Simon_2015_ICCV}
		& - & 87.9 & 89.0 & 20  \\
	Our parts
		& - & 87.4 & \textit{89.5} & 4 \\
	\bottomrule
\end{tabular}
\label{tab:part_eval}
\end{center}
\vspace{-0.6cm}
\end{table}

Second, we compare the recognition performance of our extracted parts with the one obtained with ground-truth parts.
We have chosen the CUB-200-2011 dataset for this experiment, since it is one of the few datasets that provides these annotations.
As indicated in the introduction, we also regrouped the provided ground-truth parts in more coarse but also more distinct parts, namely ``head'', ``body'', ``legs'', and ``tail''.
Compared to original ground-truth part annotations, our experiments show that these parts yield a better recognition performance (\autoref{tab:part_eval}).
This indicates again that the quality of parts is more important than the quantity.
In the same table, we compare our classification-specific part detection with the part-based approach of Simon \etal~\cite{Simon_2015_ICCV}, who have provided their extracted part locations.
Additionally, we report in the table the recognition based on the global feature only.
Best results are achieved when combining part features and the global image feature.
While using ground-truth part annotations is slightly better, we are able to achieve better recognition results than the NAC parts proposed by Simon \etal~\cite{Simon_2015_ICCV}.
Thus, our approach yields competitive recognition accuracies without relying on ground-truth part annotations, which makes it applicable in a wider range of applications where part annotations are not available.

\vspace{-.5\baselineskip}
\subsubsection{Comparison to State-of-the-Art}

Finally, we compare our proposed method with current state-of-the-art approaches on commonly used fine-grained datasets.
The results are shown in \autoref{tab:sota_results} and the mentioned baseline uses only global image features extracted from the whole image.
Furthermore, we differentiate between methods that use only the global information and part-based methods.
Besides the method of Cui \etal, which utilizes clever pretraining of the CNN weights, we compare to other methods that use sophisticated pooling methods: bilinear pooling~\cite{lin2015bilinear} and alpha-pooling~\cite{Simon19:Implicit}.
We also report recognition results and the number of used parts for other part-based approaches.
Note that none of the approaches used ground-truth part annotations, neither during training nor in the test phase.

\autoref{tab:sota_results} shows that our approach is competitive in various fine-grained applications and achieves state-of-the-art performance on the NA-Birds dataset.
For the CUB-200-2011 dataset, we outperform a lot of part-based methods even if they are using much more parts.
This highlights once again that the quality of the parts is important and that our estimated parts contain meaningful information in only four locations.


\begin{table}[t]
\caption{Comparison of our part-based approach for fine-grained recognition with various state-of-the-art methods (\textbf{bold}~=~best, \textit{italic}~=~best part-based).}
\begin{center}
\begin{tabular}{c@{\hspace{.25cm}}l@{\hspace{.5cm}}c@{\hspace{.25cm}}c@{\hspace{.25cm}}c@{\hspace{.25cm}}c@{\hspace{.25cm}}c}
	\toprule
	& & \multirow{2}{*}{\scriptsize CUB-200-2011} & \multirow{2}{*}{\scriptsize NA-Birds} & \multirow{2}{*}{\scriptsize Flowers-102} & \multirow{2}{*}{\scriptsize Stanford-Cars} & \scriptsize maximum\\
	& & & & & & \scriptsize \# of parts\\
	\midrule
	\multirow{4}{*}{\rotatebox[origin=c]{90}{\parbox[c]{2.3cm}{\underline{Global features}}}}
	& Linear SVM
		& \multirow{2}{*}{88.5} & \multirow{2}{*}{87.5} & \multirow{2}{*}{\textbf{97.8}} & \multirow{2}{*}{91.5} & \multirow{2}{*}{-} \\
    & (baseline) & & & & & \\[.4cm]
	& Lin \etal~\cite{lin2015bilinear}
		& 84.1 & - & - & 91.3 & - \\[.05cm]
	& Simon \etal~\cite{Simon19:Implicit}
		& 86.5 & - & 96.7 & 91.6 & - \\[.05cm]
	& Cui \etal~\cite{Cui_2018_CVPR_large}
		& 89.6 & 87.9 & 97.7 & \textbf{93.5} & - \\
	[.2cm]
	\multirow{8}{*}{\rotatebox[origin=c]{90}{\parbox[c]{2.9cm}{\underline{Part-based features}}}}
	& Simon \etal~\cite{Simon_2015_ICCV}
		& 81.0 & - & 95.3 & - & 20 \\
	& Krause \etal~\cite{krause2015fine}
		& 82.0 & - & - & 92.6 & 30 \\
	& Fu \etal~\cite{Fu_2017_CVPR}
		& 85.3 & - & - & 92.5 & 2 \\
	& Zhang \etal~\cite{zhang2019unsupervised}
		& 85.4 & - & - & 92.3 & 4 \\
	& Zheng \etal~\cite{zheng2017learning}
		& 86.5 & - & - & 92.8 & 5 \\
	& He \etal~\cite{he2019and}
		& 87.2 & - & - & \textit{93.3} & 15 \\
	& Ge \etal~\cite{Ge19Weakly}
		& \textbf{\textit{90.4}} & - & - & - & 10 \\
	[.15cm]
	& Our parts
		& 89.5 & \textbf{\textit{88.5}} & \textit{96.9} & 92.5 & 4 \\
	\bottomrule
\end{tabular}
\label{tab:sota_results}
\end{center}
\vspace{-0.6cm}
\end{table}

\section{Conclusion}
\label{sec:conclusion}

In this paper, we proposed a weakly supervised classification-specific part estimation approach for fine-grained visual categorization.
Unlike other part-based approaches, we estimate the part extents based on an initial classification of the whole image.
We have shown that part features extracted in a classification-specific manner result in improved categorization performance.
Furthermore, each estimated bounding box part has an implicit spatial extent that automatically determines an appropriate scale of the part.

	\newpage
	\bibliographystyle{splncs04}
	\bibliography{121-main}

\end{document}